\documentclass[runningheads]{llncs}

\usepackage{graphicx}	
\usepackage{comment}
\usepackage{amsmath,amssymb}
\usepackage{color}
\usepackage{url}
\usepackage{hyperref}

\newcommand{\acc}{\mathrm{accuracy}}

\newcommand{\TP}{\mathrm{TP}}
\newcommand{\TN}{\mathrm{TN}}
\newcommand{\FP}{\mathrm{FP}}
\newcommand{\FN}{\mathrm{FN}}
\newcommand{\IoU}{\mathrm{IoU}}
\newcommand{\mIoU}{\mathrm{mIoU}}

\newif\ifreview
\reviewtrue
\reviewfalse

\ifreview
	\usepackage{lineno}

	\linenumbers
\fi

\begin{document}


\def\SubNumber{000}

\def\GCPRTrack{Track: Computer vision systems and applications}

\title{Detecting Slag Formations with Deep Convolutional Neural Networks}

\ifreview
\else

	\author{Christian von Koch\inst{1}\orcidID{0000-0003-3222-0086
} \and
	William Anzén\inst{1}\orcidID{0000-0002-9313-3851} \and
	Max Fischer\inst{1}\orcidID{0000-0001-5882-3980} \and
	Raazesh Sainudiin\inst{1,2,3}\orcidID{0000-0003-3265-5565}}
	
	\authorrunning{C. von Koch et al.}
	
	\institute{Combient Mix AB, Stockholm, Sweden \footnote[1]{\url{https://combient.com/mix}} \and Combient Competence Centre for Data Engineering Sciences, and \and Department of Mathematics, Uppsala University, Uppsala, Sweden \footnote[2]{\url{https://math.uu.se/}}
	\email{\{christian.von.koch,william.anzen,max.fischer\}@combient.com}\\
	\email{raazesh.sainudiin@math.uu.se}}
\fi

\maketitle              

\begin{abstract} 
We investigate the ability to detect slag formations in images from inside a Grate-Kiln system furnace with two deep convolutional neural networks.  
The conditions inside the furnace cause occasional obstructions of the camera view. 
Our approach suggests dealing with this problem by introducing a convLSTM-layer in the deep convolutional neural network.  
The results show that it is possible to achieve sufficient performance to automate the decision of timely countermeasures in the industrial operational setting. 
Furthermore, the addition of the convLSTM-layer results in fewer outlying predictions and a lower running variance of the fraction of detected slag in the image time series.
\end{abstract}
%
%
%

\section{Introduction}
\label{cha:intro}
The mining industry is an ancient industry, with the first recognition of underground mining dating back to approximately 40,000 BC [5]. Historically, mining consisted of heavy labour without the aid of modern technology, but today the industry portrays a different story. Large machines and production lines are responsible for carrying out the mining process more efficiently and at scale.  
This is evident in the process of transforming raw iron ore into iron ore pellets -- small iron ore ``balls'' used as the core component in steel production.

Iron ore pellets are produced in large pelletising plants, where iron ore concentrate is mixed with bentonite, a clay mineral acting as a binder. The ore and bentonite mix is then formed into pellets in a rotating drum. At the final stage of this process, the wet iron ore pellets are pre-heated and dried in a large furnace called the Grate-Kiln system. The Grate-Kiln system \cite{pelletizing} consists of three main units: 
1.~Drying and pre-heating (Grate),
2.~Heating (Kiln)
and 3.~Cooling.

When the pellets flow between the Grate and the Kiln, a common challenge in pelletising plants is that an unwanted by-product known as slag slowly builds up over time. 
The slag gets attached to the base (slas) of the Kiln and restricts the flow of the pellets, worsening the quality of the end-product and, if it breaks loose, it can potentially damage the equipment.
Thus, it is of vital importance to track the slag build-up over time, as timely countermeasures have to be taken to remove the slag before it grows too large. 

One solution used to track the build-up is to attach a video camera positioned to record the slag inside the furnace. 
The video stream is then manually and visually analysed by employees at the plant. 
While this process enables plant operators to get an instant snapshot of the current state of the slag, the process is not only labour-intensive but also prone to inconsistencies due to individual human biases. 
Moreover, it does not provide a historical systems-level overview of how the amount of slag has changed over time. 

\begin{figure}
    \centering
    \includegraphics[scale=0.46]{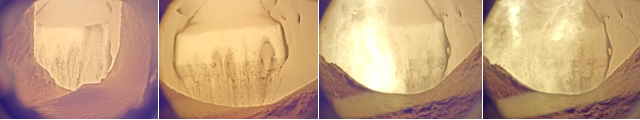}
    \caption{Sample images with varying conditions from within the Kiln. The two images to the left are occlusion-free and the two images to the right contain occlusions.}
    \label{fig:Slag-ex}
\end{figure}

Our aim is to research the possibility of measuring the amount of slag that has accumulated in the Kiln through image segmentation with deep convolutional neural networks (DCNNs). Furthermore, we want to assess the effects of utilising consecutive pairs of images in the time series of images when segmenting possibly occluded images (e.g.~flames and smoke in the third and fourth images of Fig.~\ref{fig:Slag-ex}). 
This will be done by introducing a convLSTM-layer in the DCNN. Our work contributes with the following novel findings:

\begin{enumerate}
\item{We show that it is possible to measure the amount of slag that has accumulated in a Grate-Kiln system through image segmentation with DCNNs.}
\item{We show that it is possible to reduce variance in the predicted slag fraction by using consecutive pairs of images through a convLSTM-layer in the DCNN.}
\item{We demonstrate an approach for handling images containing occlusion in the video stream from the Grate-Kiln system.  We believe that this approach can be applied to other applications where occasional obstructions of the camera view are considered problematic.}
\item{We have developed image segmentation models to predict slag formation through a scalable \cite{ZahariaXinEtAl16cacm} delta lake architecture \cite{deltalake2020} and deployed in production, thus showing that our work is not only sufficient in theory, but also brings value in practise.}
\end{enumerate}

In Sec.~\ref{method}, we describe the raw dataset and models used for conducting our experiments. 
In Sec.~\ref{cha:experiments}, we describe the first of our experiments which compares two DCNNs at the task of identifying slag formations in images. We show that it is possible to detect slag formations with both models, which to our knowledge has not been shown in the literature before. In Sec.~\ref{variance_occlusion_experiment}, we analyse how utilising temporal information might be beneficial for lowering prediction variance and handling images with occlusions in them. Finally, we summarise our results and findings in Sec.~\ref{cha:conclusions}.




\section{Related work}


Image segmentation has been used successfully in a vast number of applications such as medical image analysis, video surveillance, scene understanding and others \cite{SegmentationSurvey}. The most significant model and architecture contributions within the image segmentation field have been reviewed and evaluated by Minaee et al.~\cite{SegmentationSurvey}, whose article covers more than 100 state-of-the-art algorithms for segmentation purposes, many of which are evaluated on 
image dataset benchmarks. 

The use of image segmentation models in the field of mining is limited in the literature, but some research has been conducted on measuring the size of pellets and other mining-related particles through computer vision. Hamzeloo, et al.~used a combination of Principal Component Analysis (PCA) and neural networks to estimate particle size distributions on industrial conveyor belts \cite{SizeIronOre2014}. 
Support Vector Machines (SVMs) have been successfully used to estimate iron ore green pellet size distributions in a steel plant \cite{SizeIronOre2016} and a DCNN (U-Net architecture) has been proposed for the same purpose \cite{SizeIronOre2019}.

To our knowledge, no research has been conducted on detecting slag formation through computer vision within furnaces used in the mining industry. A reason for this might be due to the Grate-Kiln system previously being regarded as a black-box process \cite{GrateKilnBook}. Gathering data through measuring devices that can withstand temperatures above 1000$^\circ$C has been difficult historically.

When evaluating videos or sequences of images, where potential insights from temporal information can be obtained, there are multiple approaches that can be used. For example, one can concatenate the current frame with the previous \(N\) frames to a common input and process these frames with an ordinary CNN. However, Karpathy et al. \cite{Karpathy_2014_CVPR} showed that this only results in a slight improvement and that the use of a temporal model proved to be a more successful approach. Multiple authors have since then proposed the use of recurrent neural networks (RNNs) when performing image segmentation on videos or sequences of images. For example, Valipour et al.~\cite{RecurrentFCN} proposed a modification of a Fully Convolutional Network (FCN) \cite{FCN} with an added RNN unit between the encoder and decoder, achieving a higher performance on the Moving MNIST dataset \cite{MNIST} and other popular benchmarks. Yurdakul et al.~\cite{RecurrentStructures} researched different combinations of convolutional and recurrent structures, finding that an additional convLSTM cell yielded the best result. Pfeuffer et al.~\cite{convLSTMSoftmax} investigated where to place the convLSTM cell in the DCNN model architecture, finding that the largest increase in performance was achieved by placing it just before the final softmax layer. 
We use insights from such recent work on image segmentation, but with novel adaptations to our problem of slag detection above 1300 $^\circ$C.

\section{Raw dataset and models}
\label{method}

In this section, the raw dataset and the deep convolutional neural network (DCNN) models for image segmentation used in our experiments are introduced.

\subsection{Raw dataset}

All the data used in this report was supplied by LKAB (a Swedish mining company) and collected from one of their pelletising plants located in Kiruna, Sweden.
The dataset at our disposal consisted of a stream or time series of $640 \times 480$ pixel RGB images. 
Each image was captured inside the Kiln with a 10-second time lag,  and included metadata about when each image had been taken.  
The state of the furnace as well as the angle and position of the camera affect the characteristics of the images. For example, the furnace might be in its initial stages of heating causing a different lighting than when it has just been turned off. 
Furthermore, the camera sometimes had its position moved due to the vibrations within the furnace, causing the previous landscape to change.  
Lastly, the images captured sometimes contain a flame, used for heating the Kiln, with subsequent smoke.  
Both flame and smoke can occlude the view of the slag in the image. 
The images were ingested into a delta lake \cite{deltalake2020} to easily train our models.

\subsection{Models}
\subsubsection{U-Net.}
The first model implemented in this project was the U-Net \cite{Unet}. The U-Net is an early adaptation of the encoder-decoder architecture outperforming other state-of-the-art architectures of that time and field by a large margin. The U-Net produces high accuracy segmentation while only requiring a small training dataset. By utilising the feature maps of the encoder combined with the up-sampled feature maps of the decoder, localisation of features can be made in a higher resolution, thus increasing performance.

\subsubsection{PSPNet.}

The second model implemented for our experiments was the Pyramid Scene Parsing Network (PSPNet) \cite{PSPNet}. The PSPNet is a network that aims to utilise the global prior of a scene by downsampling the image into different sized sub-regions through average pooling. By doing this, both global and local information of each pixel can be exploited. 

The network uses a backbone of a Residual Network (ResNet) \cite{ResNet-101} with a dilated strategy, i.e., replacing some of its convolutional operations with dilated convolutions and a modified stride setting, as described in \cite{DilatedConv,Dilated2}. ResNets are based on the idea of a residual learning framework that allows training of substantially deeper networks than previously explored \cite{ResNet-101}. Following the backbone, the architecture uses its proposed pyramid pooling module, a concatenating step, and finally a convolutional layer to create the final predicted feature maps. 

\section{Detecting slag accumulation with deep neural networks}
\label{cha:experiments}
Here, we discuss the experiment with the aim of researching the potential of measuring slag accumulation using image segmentation with the U-Net and the PSPNet, two DCNN models. 

\subsection{Dataset}
\label{dataset-detect}
To train the DCNN models, a subset of the raw images was extracted. 
These images were manually and carefully chosen to ensure a high variance that is representative 
of the full distribution of images.
Furthermore, the images chosen did not contain any occlusions to reduce interference and noise in the training process.
Finally, the subset of images were manually labelled into four classes using the COCO-annotator tool \cite{cocoannotator}: \textit{background}, \textit{slag}, \textit{camera edge} and \textit{wall} (see Fig.~\ref{fig:ClassDistributionTrain}).
The classes were chosen based on discussions with domain experts at LKAB. \textit{Background} and \textit{slag} were needed to compute the percentage of slag stuck on the flat surface of the Kiln (i.e. the slas). \textit{Camera edge} was needed to alert local plant maintenance when the camera hole needed cleaning. \textit{Wall} simply represented the rest of the pixels.
This resulted in a dataset consisting of 530 images, with 339 used for training, 85 for validation, and 106 for testing.

\subsection{Implementation details}
\label{implementation-details}

\subsubsection{U-Net.}
The U-Net implemented in our work was built as described in the original paper, with the exception of using less filters in its layers. The number of filters used was reduced by a factor of 16 in comparison to the original architecture, e.g., the first layer of the original architecture used 64 filters and our implementation used 4 filters in the same layer.  The decision of using fewer filters in the implementation was made since no significant improvement in performance was observed by using the full-scale U-Net.	

\subsubsection{PSPNet.}
The PSPNet was built 
with a backbone of ResNet-50, a version of the ResNet model \cite{ResNet-101} with 50 convolutional layers. 
This was done by utilising the pre-built ResNet-50 model available in TensorFlow \cite{resnet-keras}, with slight modifications to suit the dilated approach of the PSPNet, as described in \cite{DilatedConv,Dilated2}.  
The weights of the backbone ResNet-50 was initialised using pre-trained weights, trained on the ImageNet dataset.

\begin{figure}
    \centering
    \includegraphics[scale=0.35]{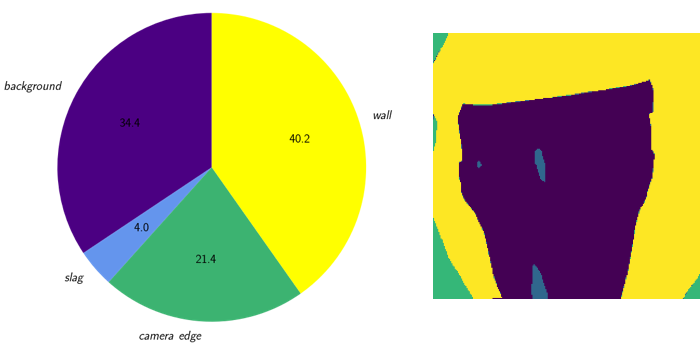}
    \caption{Visualisation of the skewed class distribution present in the dataset used for training the models in experiment 1. The classes in the image are background (purple), slag (blue), camera edge (green) and wall (yellow). (\textit{Left}): Class distribution of the data. (\textit{Right}): Example ground truth image showing the skewed distribution of pixels. }
    \label{fig:ClassDistributionTrain}
\end{figure}

\subsubsection{Loss functions and class-weighting schemes.}
When analysing the class distribution of the dataset defined in Sec.~\ref{dataset-detect}, it became evident that some classes were under-represented, causing a \textit{class imbalance problem}; see Fig.~\ref{fig:ClassDistributionTrain}. Class imbalance within a dataset has been shown to decrease performance for some classifiers, including deep neural networks, causing the model to be more biased towards the more frequently seen classes \cite{class-imbalance}. To tackle this problem, we experimented with three loss functions; the \textit{Tanimoto loss} \cite{ResUNet-a}, the \textit{dice loss} \cite{VNetDice}, and the \textit{cross-entropy loss}. We also experimented with two class-weighting schemes; the \textit{inverse square volume weighting} (ISV) and the \textit{inverse square root volume weighting} (ISRV), in comparison to no class-weighting scheme. The ISV weighting scheme was calculated by 
$w_c = 1/f_c^2$ and the ISRV weighting scheme was calculated by $w_c = 1/\sqrt{f_c}$, for each segmentation class $c$, where $f_c$ denotes the frequency of pixels labelled as class $c$ in the dataset used for training the model.  The ISV weighting scheme together with the Tanimoto loss as well as the ISRV weighting scheme together with the cross-entropy loss function have been successfully used for class imbalance problems \cite{ResUNet-a,inverse-root-two}. We found that the best performance in our trials 
was achieved by the dice loss with ISV and the cross-entropy loss with ISRV for the U-Net and the PSPNet models, respectively. For more details on ablation studies please see Sec.~5.2 of \cite{von2021detecting}.

\subsection{Evaluation metrics}
\label{sec:Metrics}

\subsubsection{Pixel Accuracy.}
The most common evaluation metric used for classification problems is $\acc$.  
For image segmentation tasks, pixel accuracy is frequently used.  It is defined as
the proportion of correctly classified pixels, as follows:

\begin{equation}
    \acc := \dfrac{\TP+\TN}{\TP+\FP+\TN+\FN}\enspace,
\end{equation}

\noindent where $\TP$, $\FP$, $\TN$ and $\FN$ refer to the \textit{true positives}, \textit{false positives}, \textit{true negatives} and \textit{false negatives}, respectively.  

\subsubsection{Intersection over Union.}
\noindent One of the most commonly used metrics for image segmentation is the \textit{intersection over union}, denoted by $\IoU$ \cite{SegmentationSurvey}. 
This metric measures the overlap of pixel classifications, making it less sensitive to class imbalances within the image, in comparison to $\acc$. 
The equation for $\IoU$ for each class $c$ is defined below, where $A$ denotes the predicted segmentation map over the classified pixels and $B$ denotes the ground truth segmentation map. 

\begin{equation}
\label{eqn:IoU}
    \IoU_c := \dfrac{|A \cap B|}{|A \cup B |}
\end{equation}

\noindent The $\IoU$ of the class labelled {\em slag} was used for validation since this was the class of interest in this problem.  

\subsubsection{Mean Intersection over Union.}
\noindent The \textit{mean intersection over union} metric, denoted by $\mIoU$, averages over the $\IoU$ metric for each of the classes and is defined as follows:

\begin{equation}
	\mIoU := \frac{1}{|\mathbb{C}|}\sum_{c\in \mathbb{C}} \IoU_c \enspace,
\end{equation}

\noindent where $|\mathbb{C}|$ is the number of classes in set $\mathbb{C}$ and $\IoU_c$ is the $\IoU$ for class $c$. 

\subsection{Results and analysis}

\def\arraystretch{1.5}
\setlength{\tabcolsep}{0.5em}
\begin{table}[hbtp]
\centering
\caption{Class-wise $\IoU_c$ scores of the slag detection experiment with $\mIoU$ and $\acc$ calculated over all classes. All metrics are computed on the test set.}\label{slag-detect}
\begin{tabular}{l|l|l|l|l}
\textrm{Model} & \textrm{Class} $c$ & $\IoU_c$(\%) & $\mIoU$(\%) & $\acc$(\%) \\
\hline
\textrm{PSPNet} & 
\begin{tabular}[c]{@{}l@{}}{\itshape background}\\ {\itshape slag}\\ {\itshape camera edge}\\ {\itshape wall}
\end{tabular} & 
\begin{tabular}[c]{@{}l@{}}94.06\\ 65.33\\ 94.63\\ 96.85
\end{tabular} & 87.72 & 97.32 \\
\hline
\textrm{U-Net} & 
\begin{tabular}[c]{@{}l@{}}{\itshape background}\\ {\itshape slag}\\ {\itshape camera edge}\\ {\itshape wall}
\end{tabular} & 
\begin{tabular}[c]{@{}l@{}}92.54\\ 62.67\\ 91.61\\ 94.67
\end{tabular} & 85.37 & 96.38 \\
\hline
\end{tabular}
\end{table}

\noindent The results of this experiment are presented in \autoref{slag-detect}. 
The results show that both models are able to segment the images in the test set, especially for the more frequent class labels: {\em background}, {\em camera edge} and {\em wall}. Both the U-Net and PSPNet are also able to segment slag at an $\IoU$ of 62.67\% and 65.33\%, respectively,  thus showing that it is possible to utilise image segmentation when detecting slag. This was also visually verified by domain experts at LKAB,  verifying that the achieved $\IoU$ of slag was sufficient in an industry setting.

\begin{figure}[htbp]
    \centering
    \includegraphics[scale=0.21]{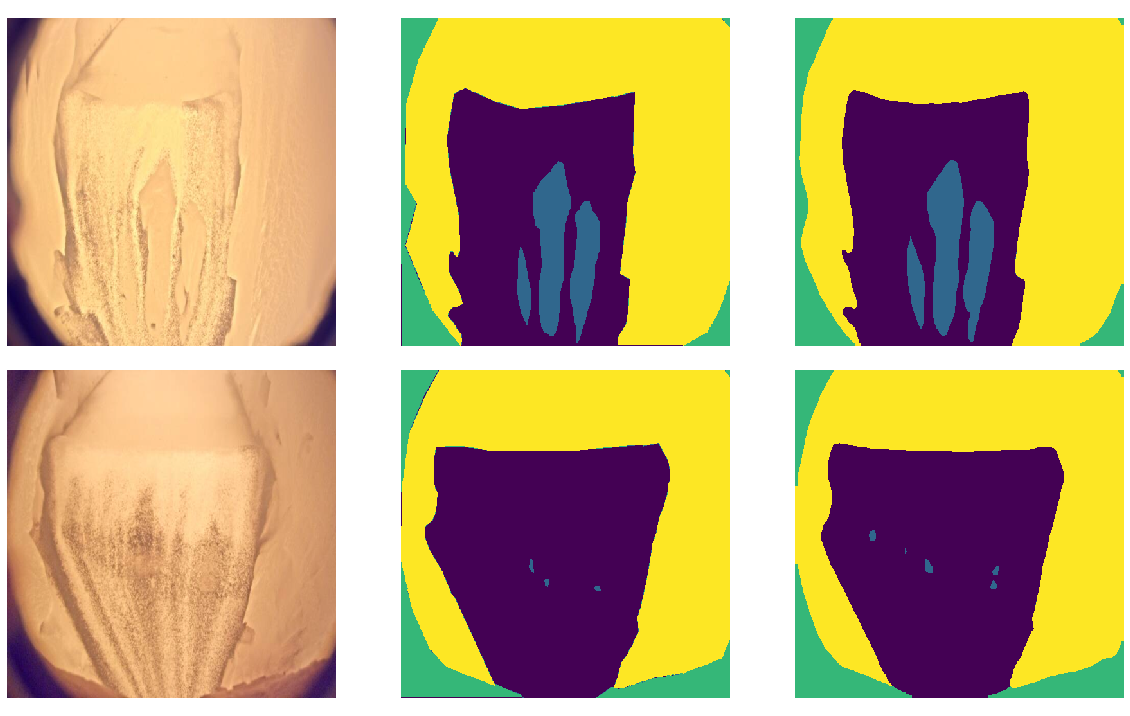}
    \caption{Two examples of predictions yielded by the PSPNet with weighted loss. (\textit{Top}): The best prediction, yielding $86.8$\% $\IoU$ of \textit{slag}. (\textit{Bottom}): The worst prediction, yielding $8.4$\% $\IoU$ of \textit{slag}. (\textit{Left}): The input images. (\textit{Middle}): The ground truth images. (\textit{Right}): The predictions made by the PSPNet.}
    \label{fig:pspnet_prediction}
\end{figure}

The lower $\IoU$ for the class {\em slag} in comparison to the other classes we believe is due to the class imbalance problem, mentioned in Sec.~\ref{implementation-details}. Another reason could be an effect of the manual labelling process, as it is sometimes difficult for the human eye to detect the slag islands due to their small size and sometimes indistinguishable edges. This is evident in the worst prediction yielded by the PSPNet visualised in Fig.~\ref{fig:pspnet_prediction}, which also shows that even though the prediction of the model yielded a low $\IoU$ score in this case, the \textit{amount} of slag predicted can be fairly accurate.

Although both of the models showed potential in segmenting the images, the superiority of the PSPNet over the U-Net in this particular context is evident from looking at the results in \autoref{slag-detect}, as it outperforms the U-Net in all of the metrics measured. 
Fig.~\ref{fig:pspnet_prediction} displays some example predictions made by the PSPNet which we believe shows, together with the results presented in \autoref{slag-detect}, that it is possible to measure the amount of slag that has accumulated in a Grate-Kiln system through image segmentation with a DCNN. We will in the next experiment look into whether the better performing PSPNet is consistent in its predictions by analysing its performance on unlabelled images over a longer time-frame. 

\section{Variance and occlusion experiment}
\label{variance_occlusion_experiment}
Here, we present the experiment with the aim of researching the performance in a productionised setting. It includes analysing the prediction variance and the ability to deal with occluded images hindering the view of the camera.

\subsection{Datasets}

\subsubsection{Training dataset.}
\label{training-dataset}
The training dataset, used in Sec.~\ref{cha:experiments}, was extended by adding the subsequently captured images, captured with a 10 second time lag.  This resulted in a time series dataset containing consecutive pairs of images. 
As the images captured in the Kiln sometimes include flames and subsequent smoke obstructing the view, the resulting dataset only contained 420 pairs of images, i.e., 110 out of the 530 subsequent images were discarded due to the presence of an occlusion. The newly added images were labelled in the same manner as was described in the first experiment; see Sec.~\ref{dataset-detect}. Out of the 420 pairs of images, 269 pairs were used for training, 67 used for validation, and 84 used for testing. After testing the performance of the models on unseen data, we retrained the models prior to evaluation on the following \textit{evaluation datasets}, using 336 pairs for training and 84 for validation.

\subsubsection{Evaluation datasets.}
\label{evaluation-datasets}
To evaluate the robustness of the models in a productionised setting, raw consecutive images taken in sequence from two separate days were extracted into two datasets, containing 7058 and 8632 images respectively. The two days were chosen because of their different characteristics. One of the days contained images that showed a slow and gradual slag build-up, whereas the other day contained images where it was known beforehand that slag had been removed throughout the day, resulting in a varying slag build-up. These images were not labelled. 

\subsection{Implementation details}
\subsubsection{Occlusion Discriminator.}
\label{occlusion-discriminator}
In order to evaluate the DCNN models on the unlabelled evaluation datasets, the images input to the models needed to be occlusion-free. To replicate a production setting and to reduce the manual work of removing images containing occlusions, an \textit{occlusion discriminator} model was implemented and trained. The model was a DCNN classifier of three convolutional blocks, each block containing two convolutional layers each of which was followed by a batch normalisation and ReLU activation functions. 
Each convolutional block was then followed by a max pooling and dropout layer. 
Finally, a dense layer was followed by a sigmoid function, yielding binary predictions on whether the image contains occlusions or not. After training the model, it achieved a precision and recall score of 99.7\% and 83.7\% respectively.
This model was used to filter the evaluation datasets.

\subsubsection{Two PSPNet models. }
The best performing hyperparameters of the PSPNet from the previous experiment, described in Sec.~\ref{cha:experiments}, was used in this experiment. It was trained using the cross-entropy loss with the ISRV class-weighting scheme, and is referred to as the PSPNet. The PSPNet was evaluated in comparison to a modified version of the same PSPNet model, referred to as the PSPNet-LSTM.  The PSPNet-LSTM was modified by replacing the final convolutional layer with a convLSTM-layer with two cells, with shared weights, to enable the model to process pairs of images. Furthermore, the layers prior to the convLSTM-layer were implemented to ensure that both of the input images were processed by the same weights. 

\begin{figure}[htbp]
    \centering
    \includegraphics[scale=0.145]{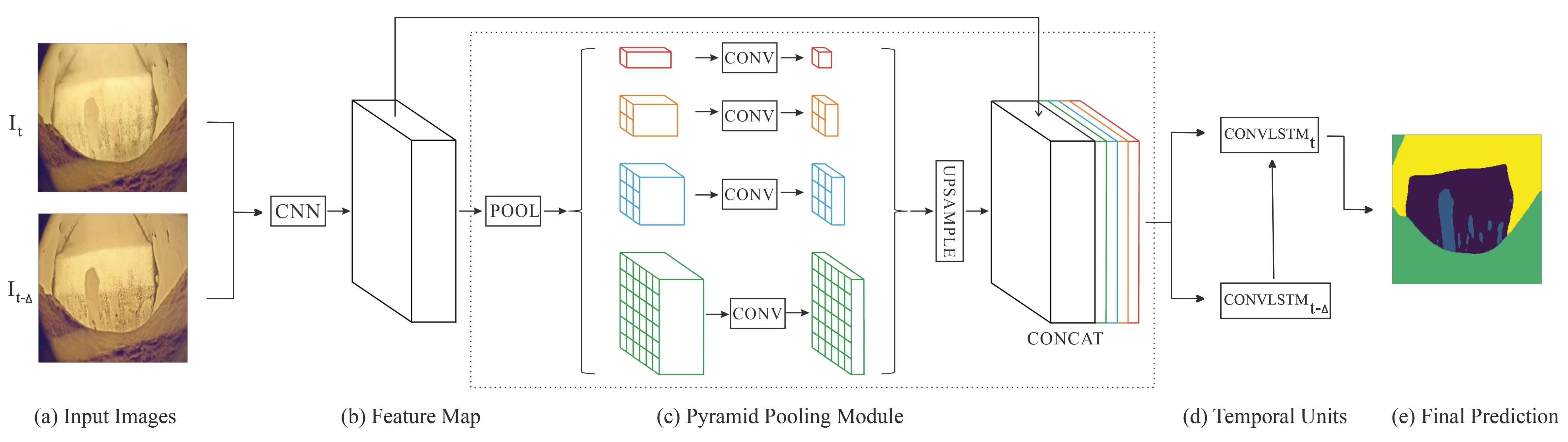}
	\caption[Visualisation of the PSPNet-LSTM, our modified PSPNet, with convLSTM cells]{Visualisation of the PSPNet-LSTM, our modified PSPNet with convLSTM cells. The images $I_t$ and $I_{t-\Delta}$, seen in (a), are both processed by the same weights in the network.  The feature maps produced from $I_t$ are fed into the convLSTM unit at time step $t$, and the feature maps produced from $I_{t-\Delta}$ are fed into the convLSTM unit at time step $t-\Delta$. Finally, a segmentation map prediction of the image $I_t$ is made.}
    \label{fig:PSP_LSTM}
\end{figure}

The convLSTM-layer utilised information from a sequence of images. 
When predicting a segmentation mask of an image $I_t$ of time step $t$, it used information from both $I_t$ and the hidden state of the previous LSTM cell utilising information from $I_{t-\Delta}$, with $\Delta$ set to 10 seconds in our experiments. 
The output of the PSPNet-LSTM was a segmentation mask prediction of the image $I_t$. See a visualisation of the model in Fig.~\ref{fig:PSP_LSTM}. In summary, these modifications enabled the PSPNet-LSTM to train on pairwise images.

To be able to compare the performance of the two models, the PSPNet model was retrained using the images (only $I_t$) defined in Sec.~\ref{training-dataset}. The weights of PSPNet-LSTM were then initialised using the frozen pre-trained weights from the PSPNet. By using the same weights in both models and only training the additional convLSTM-layer in PSPNet-LSTM (using both $I_t$ and $I_{t-\Delta}$), a comparison between the predictive ability of the two models on the images, with and without temporal information, could be made.

\subsection{Evaluation Metrics}


\subsubsection{IoU, fraction of classified slag and running variance.}
The $\IoU$ of the class {\em slag}, defined in \eqref{eqn:IoU}, was used for validation and testing, when training the PSPNet.  
Since the purpose of this experiment was to evaluate the stability of the models in a productionised setting, raw images from two days were used for evaluation, defined in Sec.~\ref{evaluation-datasets}. These images were not labelled, thus the $\IoU$ could not be used.  As a measurement of stability over time, the fraction of classified slag was used for evaluation:

\begin{equation}
	\mathfrak{f}_t = \frac{n^{(t)}_s}{H \times W}
\end{equation}

\noindent where $n^{(t)}_s$ is the number of pixels classified as slag and $H$ and $W$ represent the dimensions of the image at time $t$. 

The running variance was calculated for each measurement of the fraction of classified slag at time $t$, i.e., $\mathfrak{f}_t$, over a fixed window of size $k$, prior to $t$. We set $k$ to 60 amounting to 600 seconds or 10 minutes of the natural operational window for counter-measures.. 

\begin{equation}
	\mathrm{Variance}(\mathfrak{f}_{t-k+1} +\mathfrak{f}_{t-k+2} + ... + \mathfrak{f}_{t})
\end{equation}

\subsection{Results and analysis}

When comparing the best performing PSPNet with the best performing PSPNet-LSTM, evaluated on the unseen test data held out during training (Sec.~\ref{training-dataset}), the PSPNet outperformed the PSPNet-LSTM with a small margin in terms of the $\IoU$ for the class {\em slag}, yielding $62.39$\% and $61.76$\% respectively.  
The models were then evaluated on the two days of images to measure their robustness in a productionised setting by using the running variance of slag fraction --- a domain-specific metric that can highlight discrepancies in performance by manual examination before operational countermeasures.

\begin{figure}[h]
    \centering
    \includegraphics[scale=0.42]{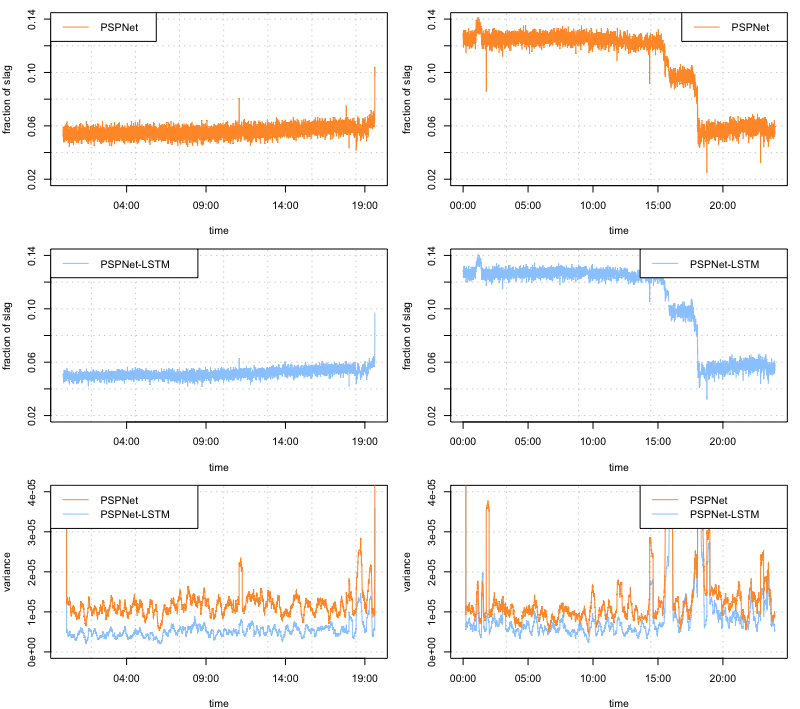}
    \caption[Predicted fraction of slag and running variance on the evaluation dataset]{Predicted fraction of slag and running variance on the evaluation dataset filtered by the occlusion discriminator.}
    \label{fig:result_exp_two}
\end{figure}

Fig.~\ref{fig:result_exp_two} shows the predicted fraction of slag on the raw evaluation dataset (two full days of unlabelled images) and the corresponding running variance. The figures show that the predictions yielded by the PSPNet-LSTM has fewer outliers and is more consistent compared to the PSPNet predictions. Furthermore, the running variance is consistently lower for the PSPNet-LSTM during the period when the slag is slowly accumulating. A low and stable running variance is preferred since it should resemble the slow process of slag build-up.

\begin{figure}[h]
    \centering
    \includegraphics[scale=0.30]{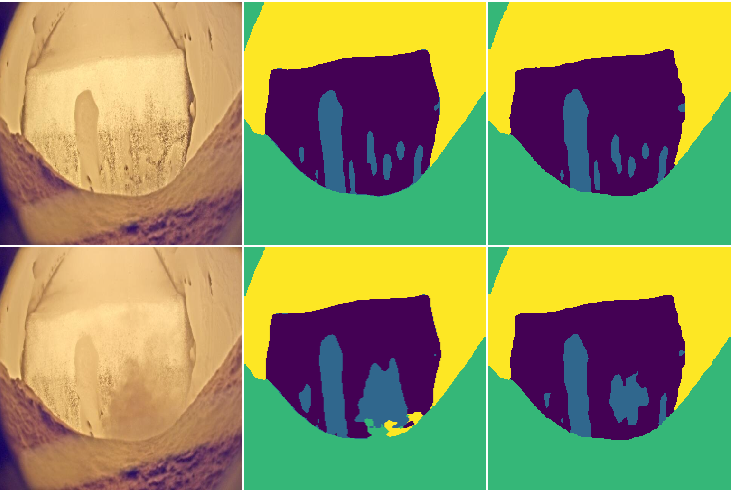}
    \caption[Occlusion that bypassed the occlusion discriminator with predictions]{(Top row): Image, $I_{t-\Delta}$, with no visible occlusion. (Bottom row): Consecutive image, $I_t$, with a visible occlusion that passed through the occlusion discriminator.  (\textit{Left}): The input image.  (\textit{Middle}): The prediction made by the PSPNet. (\textit{Right}): The prediction made by the PSPNet-LSTM.}
    \label{fig:Outliers}
\end{figure}

We believe that the reason for the more stable predictions made by the PSPNet-LSTM is its ability to produce an adequate segmentation map surrounding one occluded image where it can partly disregard an occlusion obstructing the view. Even with an occlusion discriminator in production filtering out most occlusions, false negatives are likely to occur when used over longer periods of time.  Fig.~\ref{fig:Outliers} shows an image that was wrongly classified and passed through the trained occlusion discriminator filter (Sec.~\ref{occlusion-discriminator}).  Looking at the predictions made by the two models, combined with the fewer outlying predictions of the PSPNet-LSTM in Fig.~\ref{fig:result_exp_two}, we arrive at the conclusion that the PSPNet-LSTM is more robust in its predictions and manages to predict slag regions behind the smoke, which obstructs the view, even though it has not been trained on images containing occlusions. 

\section{Conclusions}
\label{cha:conclusions}

In this paper, we have researched the possibilities of measuring slag accumulation with the use of two DCNNs, the U-Net and the PSPNet. Furthermore, we have looked into how to handle images containing occlusions and proposed a novel approach to tackle this problem. 
We have shown that it is possible to measure slag accumulation in a Grate-Kiln system using a DCNN. 
The best performing model, the PSPNet, yielded an $\IoU$ of \textit{slag} of $65.33$\%, an $\mIoU$ of $87.72$\% and an accuracy of $97.32$\%, all measured on test data. 
Finally, we have shown that a DCNN with an additional convLSTM-layer with two cells can increase the stability and lower the variance of the predictions by exploiting temporal information in consecutive images, thus making it more suitable for a production environment. 

Based on interviews with domain experts at LKAB, having access to continuous measurement of slag formation will support their effort in putting in timely countermeasures to remove slag, as well as getting a better understanding of when enough slag has been removed. 
Additionally, having a historical and systemic view of slag formation will enable LKAB to correlate the slag build-up with other parts of the Grate-Kiln system -- an important step towards understanding the root cause of the slag build-up.
Such a view can be obtained by LKAB due to the raw images over multiple years being made readily available in a delta lake house to quickly build new AI models over GPU-enabled Apache Spark \cite{ZahariaXinEtAl16cacm} clusters, and by integrating image data with other data sources.  

Future work will focus on how to further enhance the ability of the model to deal with occluded images. 
We believe that this could be done by utilising longer sequences of images when making predictions.  This will increase the probability of having multiple consecutive images without any occlusion, thus hopefully aiding the model in making accurate predictions. Moreover, our image delta lake architecture, over long time scales, allows to take a systems-level approach to the problem by incorporating other dependent variables.

\subsubsection*{Acknowledgements}
This research was partially supported by the Wallenberg AI, Autonomous Systems and Software 
Program funded by Knut and Alice Wallenberg Foundation and Databricks University Alliance with AWS credits.
We thank Gustav Häger and Michael Felsberg at Computer Vision Laboratory, Department of Electrical Engineering, 
Linköping University for their support. 
Many thanks to Håkan Tyni, Peter Alex, David Björnström and the rest at LKAB for answering all of our questions 
and supplying us with the raw data.
We are grateful to Ammar Aldhahyani for the custom illustration in Fig.~\ref{fig:PSP_LSTM} with the kind permission of Zhao to modify the original image \cite{PSPNet}.
%
%
%
\bibliographystyle{splncs04}
\bibliography{myrefs}

\end{document}